\documentclass[letterpaper, 10 pt, journal, twoside]{ieeetran}

\IEEEoverridecommandlockouts                              

\usepackage{cite}
\usepackage{hyperref}
\usepackage{graphicx}
\usepackage{sidecap}
\usepackage{wrapfig}
\usepackage{subfig}
\usepackage{amssymb,amsmath,physics}
\usepackage{array,booktabs,arydshln}
\usepackage[cal=boondox,scr=boondoxo]{mathalfa}
\usepackage{xcolor}
\usepackage{algorithm}
\usepackage{algorithmic}
\definecolor{green}{RGB}{3,112,15}
\definecolor{yellow}{RGB}{255,140,0}

\title{Toward Agile Maneuvers in Highly Constrained Spaces: Learning from Hallucination}

\author{Xuesu Xiao$^{1*}$, Bo Liu$^{1*}$, Garrett Warnell$^{2}$, and Peter Stone$^{1, 3}$
\thanks{Manuscript received: October, 8, 2020; Revised December, 20, 2020;
Accepted January, 16, 2021.}
\thanks{This paper was recommended for publication by Editor Nancy Amato upon
evaluation of the Associate Editor and Reviewers' comments. 
}
\thanks{$^{*}$Equally contributing authors}
\thanks{$^{1}$Xuesu Xiao, Bo Liu, and Peter Stone are with Department of Computer Science, The University of Texas at Austin, Austin, TX 78712 {\tt\small \{xiao, bliu, pstone\}@cs.utexas.edu}}
\thanks{$^{2}$Garrett Warnell is with the Computational and Information Sciences Directorate, Army Research Laboratory, Adelphi, MD 20783 {\tt\small garrett.a.warnell.civ@mail.mil}}%
\thanks{$^{3}$Peter Stone is with Sony AI}%
\thanks{Digital Object Identifier (DOI): see top of this page.}
}

\begin{document}
\maketitle

\begin{abstract}
While classical approaches to autonomous robot navigation currently enable operation in certain environments, they break down in  tightly constrained spaces, e.g., where the robot needs to engage in agile maneuvers to squeeze between obstacles.
Recent machine learning techniques have the potential to address this shortcoming, but existing approaches require vast amounts of navigation experience for training, during which the robot must operate in close proximity to obstacles and risk collision.
In this paper, we propose to side-step this requirement by introducing a new machine learning paradigm for autonomous navigation called {\em learning from hallucination} (LfH), which can use training data collected in {\em completely safe} environments to compute navigation controllers that result in fast, smooth, and safe navigation in highly constrained environments.
Our experimental results show that the proposed LfH system outperforms three autonomous navigation baselines on a real robot and generalizes well to unseen environments, including those based on both classical and machine learning techniques. 
\end{abstract}

\section{INTRODUCTION}
\label{sec::intro}

\IEEEPARstart{A}{utonomous}
navigation in complex environments is an essential capability of intelligent mobile robots, and decades of robotics research has been devoted to developing autonomous systems that can navigate mobile robots in a collision-free manner in certain environments \cite{fox1997dynamic}. However, when facing highly constrained spaces that are barely larger than the robot, it is difficult for these conventional approaches to produce feasible motion without requiring so much computation that the robot needs to slow down or even stop.

Recently, machine learning approaches have also been used to successfully move robots from one point to another \cite{pfeiffer2017perception}.
Those approaches, based on techniques such as Reinforcement Learning (RL) and Imitation Learning (IL), have enabled new capabilities beyond those provided by classical navigation, such as terrain-based \cite{wigness2018robot} and social \cite{chen2017socially} navigation.
Their initial success indicates a strong potential for learning-based methods to complement --- and possibly to improve upon --- classical approaches.
However, most machine learning techniques require a large amount of training data before they can generalize to unseen environments. Furthermore, these approaches typically cannot provide verifiable guarantees that the robot will not collide with obstacles while navigating to its goal.
While these shortcomings may not prove detrimental when applying machine learning to mobile robot navigation in relatively simple environments, their effects become disastrous in highly constrained spaces.
In such environments, RL methods --- which typically rely on random exploration --- are unlikely to quickly find safe controllers, especially without catastrophic failures during training.
IL methods are also unlikely to succeed due to the challenge of gathering demonstration data since highly constrained environments are typically difficult to navigate even for humans.
In short, most existing machine learning paradigms for autonomous navigation lack both (1) the ability to generate sufficient training data for learning to navigate in highly constrained spaces and (2) safety assurances to prevent collisions.

\begin{figure}[]
\centering
\includegraphics[width=1\columnwidth]{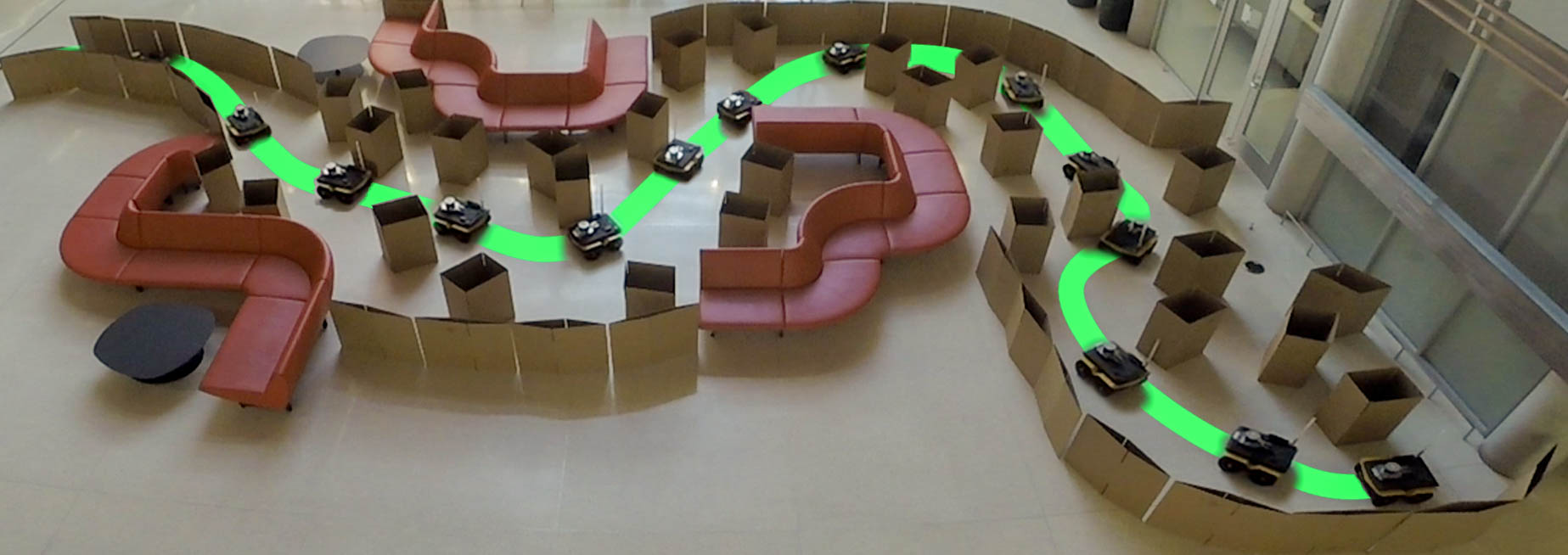}
\caption{LfH navigates in a highly constrained obstacle course. }
\label{fig::gdc}
\end{figure}

In this paper, we introduce a novel machine learning paradigm for navigation, Learning from Hallucination (LfH), that addresses the shortcomings above and enables safe, fast, and smooth navigation in highly constrained spaces. To address challenge (1), i.e. data insufficiency, we introduce a self-supervised neural controller which can collect training data in an obstacle-free environment using a randomly-exploring policy. After performing various collision-free maneuvers, highly constrained configuration spaces that allow the same effective maneuvers are synthetically projected onto the recorded perceptual data so that the machine learner can be provided with training data as if the robot had been moving in those constrained spaces.
We refer to this process of modifying the robot's real perception as {\em hallucination}.
Thanks to the inherent safety of navigating in an obstacle-free training environment, the robot can automatically generate a large amount of training data with minimal or no human supervision. In order to generalize to unseen deployment environments (e.g. environments that are less constrained), the robot's perceptual stream is also hallucinated during runtime, using whichever perceptual stream is more constrained (between real and hallucinated) as input. To address challenge (2) regarding safety, we leverage the capabilities of classical navigation approaches: the robot assesses safety at runtime using classical techniques from model predictive control, and adjusts its motion by modulating its speed and aborting unsafe plans.
LfH is fully implemented on a physical robot, and we show that it can achieve safer, faster, and smoother navigation compared to three classical and learning baselines in a highly constrained environment (Figure \ref{fig::gdc}).

This paper makes three main contributions. (1) From the motion planning perspective, LfH is a novel data-driven technique that enables safe, fast, and smooth maneuvers in previously unseen highly constrained spaces. (2) From the machine learning perspective, LfH is a novel, self-supervised learning technique that collects data offline in an obstacle-free environment and hallucinates the most constrained configuration space during training for better sample efficiency. (3) We implement LfH as an end-to-end local planner for navigation which modulates motion with explicit safety estimation and provide empirical evidence of its efficacy. Combined with other conventional navigation components, our implementation achieves safer, faster, and smoother navigation in (highly constrained) unseen environments without extensive engineering, training, or expert demonstrations, compared to classical and learning baselines. 

\section{RELATED WORK}
\label{sec::related}
This section summarizes related work from the robotics community that has sought to address autonomous navigation in highly constrained spaces. We also review the literature from the machine learning community that has considered the general problem of mobile robot navigation.

\subsection{Classical Navigation}
Given a global path from a high-level global planner, such as Dijkstra's algorithm \cite{dijkstra1959note}, A* \cite{Hart1968} or D* \cite{ferguson2006using}, classical mobile robot navigation systems aim to compute fine-grained motion commands to drive the robot along the global path while observing kinodynamic constraints and avoiding obstacles. Minguez and Montano \cite{minguez2004nearness} proposed a sophisticated rule-based Nearness Diagram approach to enable collision avoidance in very dense, cluttered, and complex scenarios and applied it on a simple holonomic robot. But for differential drive robots with non-trivial kinodynamics, researchers have relied heavily on sampling based techniques \cite{kavraki1996probabilistic, kuffner2000rrt}: Fox {\em et al.} \cite{fox1997dynamic} generated velocity samples achievable by the robot's physical acceleration limit, and found the optimal sample according to some scoring function to move the robot towards a local goal, along a global path, and away from obstacles.  Howard {\em et al.} \cite{howard2008state} sampled in the robot's state space instead of its action space, and subsequently generated motion trajectories for the sampled states. Given a highly constrained environment, the required sampling density has to increase so that a feasible motion command can be computed. Recently, Xiao {\em et al.} \cite{xiao2020appld} established that, in constrained environments, robots oftentimes need to trade off between high computational requirements for increased sampling and reduced maximum speed in order to successfully navigate. Their APPLD algorithm manages this trade off.  In contrast, LfH aims to enable safe, fast, and smooth navigation in places where the obstacle clearance is only slightly larger than the robot footprint, with limited computational requirements and without expert engineering, or human demonstrations. 

\subsection{Learned Navigation}
A flurry of recent research activity has proposed several new approaches that apply machine learning techniques to the navigation task \cite{pfeiffer2017perception, xiao2020appld, pfeiffer2018reinforced, tai2017virtual, zeng2019navigation, wang2018learning, xie2018learning, chiang2019learning, gao2017intention, zhu2017target, wang2019dual}.
While these approaches are distinct in several ways (e.g., the particular way in which the navigation problem is formulated, the specific sensor data used, etc.), the machine learning paradigm employed is typically either reinforcement learning or imitation learning.
Many approaches based on RL~\cite{sutton2018reinforcement} rely on hand-crafted reward functions for learners to discover effective navigation policies through self-generated experience.
Approaches based on IL \cite{argall2009survey, hussein2017imitation}, on the other hand, use demonstrations of effective navigation behaviors provided by other agents (e.g., humans) to learn policies that produce behaviors similar to what was demonstrated.
Both paradigms have been successfully applied to the navigation task in certain scenarios, and have even enabled new navigation capabilities beyond what is typically possible with classical autonomous navigation approaches, e.g., terrain-based \cite{wigness2018robot} and social navigation \cite{chen2017socially}.
However, these approaches each impose substantial requirements at training time: RL-based techniques rely on large amounts of training experience gathered using a typically-random exploration policy within an environment similar to that in which the robot will be deployed, and IL-based techniques require a demonstration from the same type of environment. 
Even after fulfilling these requirements, 
the learned planners were only deployed in relatively open spaces, such as hallways and race tracks. 
Unfortunately, neither of these requirements is easily satisfied in the challenging, highly constrained environments we study here: random exploration policies are too dangerous, and it is often difficult for an agent --- artificial or human --- to provide an expert-level demonstration.
In contrast, the approach we present here utilizes imitation learning, but modifies the paradigm such that it can train using arbitrary demonstrations gathered from a different, safer environment, without any trial-and-error or expert demonstrations. Furthermore, compared to hours of training time and millions of training data/steps (as in, e.g., \cite{pfeiffer2017perception}), LfH learns an entire planner within five minutes from only thousands of data points. 

\section{LEARNING FROM HALLUCINATION}
\label{sec::method}
We now describe the proposed technique, Learning from Hallucination, which can circumvent the difficulties of using traditional planning and learning approaches in highly constrained workspaces. 
In Section \ref{sec::method_motion_planning}, we formulate the classical motion planning problem in such a way that it can be easily understood from the machine learning perspective.
In Section 3.2, we propose to solve the reformulated motion planning problem with a new machine learning approach that addresses the inefficiencies of existing methods through a technique that we call hallucination. In Section \ref{sec::method_both}, we describe the technique to ensure safe navigation in challenging spaces by incorporating both classical and learning techniques to adapt to motion uncertainties. 

\subsection{Motion Planning Formulation}
\label{sec::method_motion_planning}
In robotics, motion planning is formulated in configuration space (C-space) \cite{latombe2012robot}. Given a particular mobile robot, the robot's C-space $C$ represents the universe of all its possible configurations. Given a particular environment, the C-space can be decomposed as $C = C_{obst} \cup C_{free}$, where $C_{obst} \in \mathcal{C}_{obst}$ is the unreachable set of configurations due to obstacles, nonholonomic constraints, velocity bounds, etc., and $C_{free}$ is the set of reachable configurations. Let $u\in\mathcal{U}$ be a low-level action available to the robot (e.g., commanded linear and angular velocity $\left( v, \omega \right)$), and let a plan $p \in \mathcal{P}$ be a sequence of such actions $\{u_i~|~1\leq i \leq t\}$, where $\mathcal{P}$ is the space of all plans over time horizon $t$. Then, using the notation above, the task of designing a motion planner is that of finding an optimal function $f(\cdot)$ that can be used to produce plans $p=f(C_{obst}~|~c_c, c_g)$ that result in the robot moving from the robot's current configuration $c_c$ to a specified goal configuration $c_g$ without intersecting $C_{obst}$, while observing robot motion constraints and optimizing a particular cost function (e.g. distance, clearance, energy, and combinations thereof).

To address kinodynamics constraints, lack of explicit representation of $C_{obst}$ and $C_{free}$, high dimensionality, etc., sampling-based techniques are typically used to approximate $f(\cdot)$: Probabilistic Roadmap (PRM) \cite{kavraki1996probabilistic} samples in $C_{free}$ and assumes finding $p$ to connect two consecutive configurations $c_{n-1}$ and $c_{n}$ without entering $C_{obst}$ is trivial, e.g.~using a straight line. 
Rapidly-exploring Random Tree (RRT) \cite{kuffner2000rrt} uses kinodynamic models to push samples in $C_{free}$ towards $c_g$. 
Dynamic Window Approach (DWA) \cite{fox1997dynamic} directly generates samples in action space $\mathcal{U}$ and finds the best sample to move the robot towards $c_g$ in $C_{free}$. 
However, in the case of highly constrained spaces, a large sampling density is necessary. For example, PRM requires many samples to ensure the possibility of finding valid local connections between samples without entering $C_{obst}$. RRT requires many samples to efficiently progress in $C_{free}$ toward $c_g$. DWA requires many samples to simply generate a viable action toward $c_g$ while keeping the robot configuration in $C_{free}$. Meanwhile, the output of sampling is often not a smooth trajectory, thus requiring post-processing. However, smoothing in highly constrained spaces may be difficult without the path entering $C_{obst}$. 



\begin{figure}
\centering
\subfloat[]{\includegraphics[width=0.237\columnwidth]{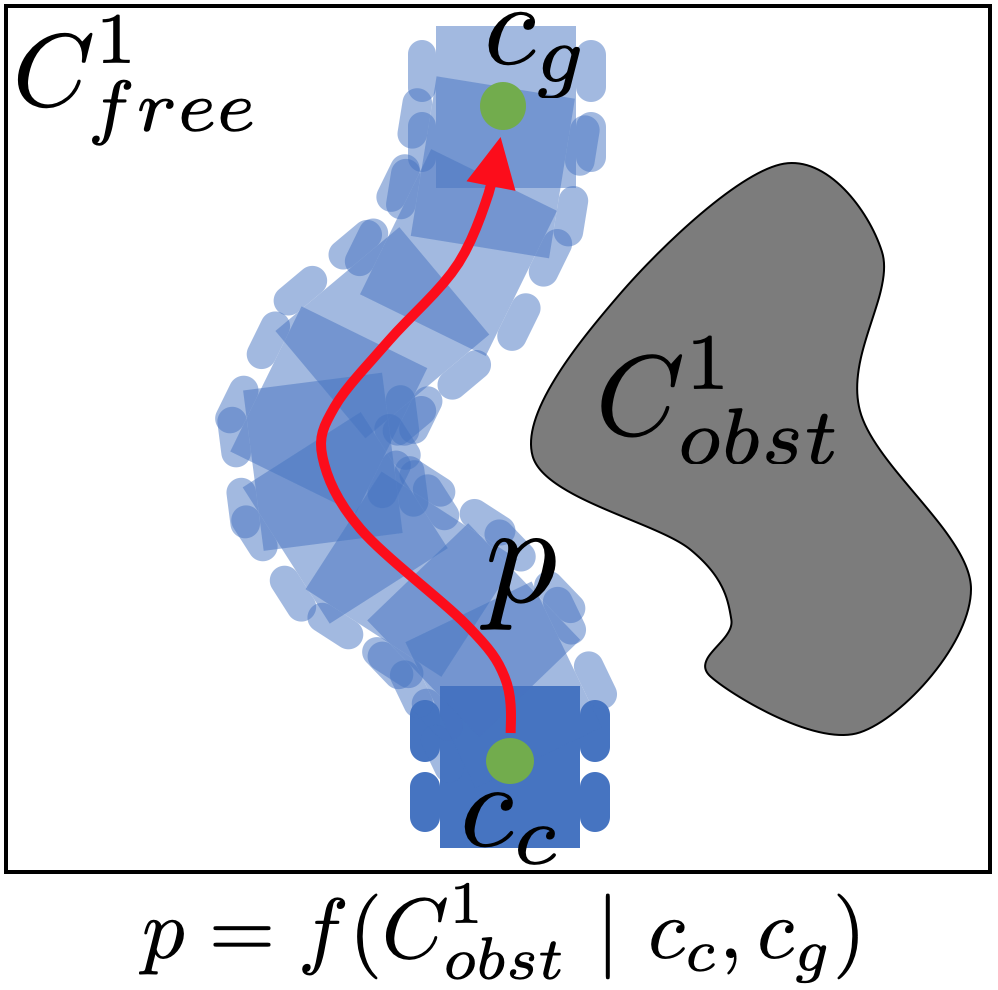}%
\label{fig::h1}}
\hspace{1pt}
\subfloat[]{\includegraphics[width=0.237\columnwidth]{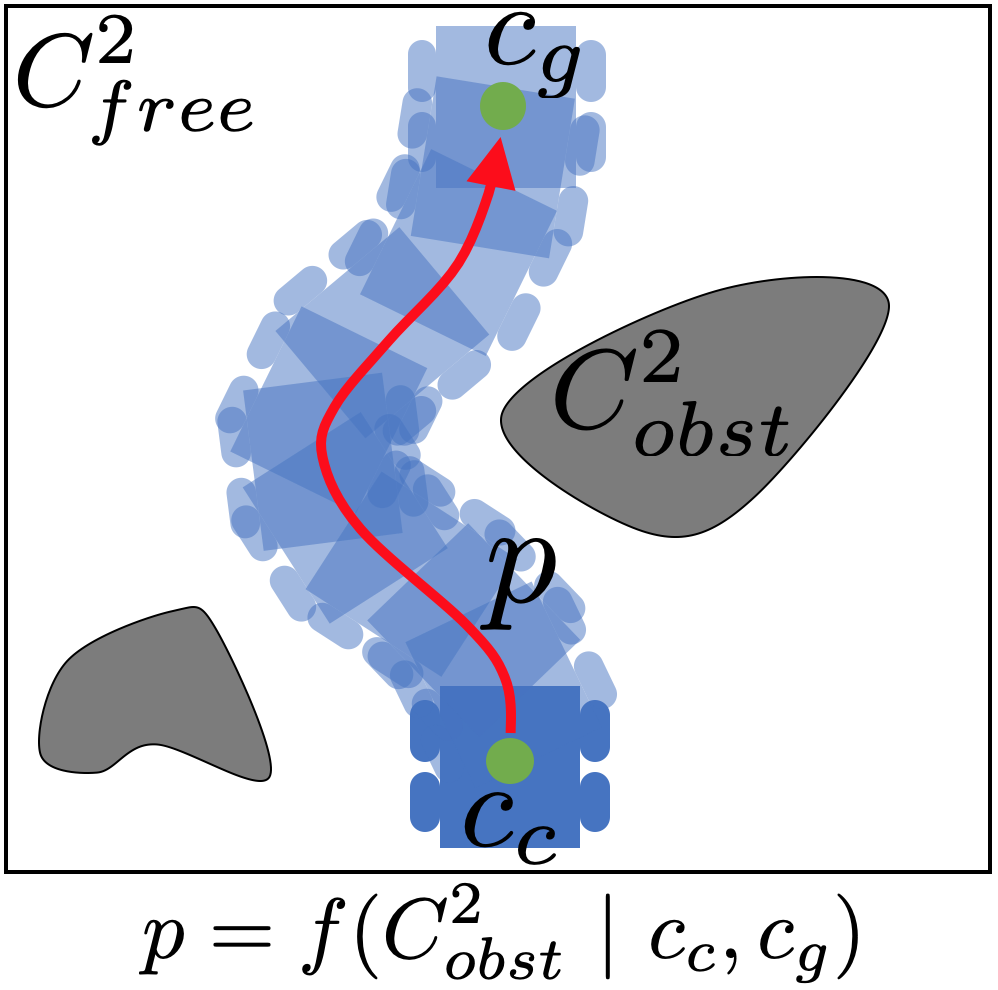}%
\label{fig::h2}}
\hspace{1pt}
\subfloat[]{\includegraphics[width=0.237\columnwidth]{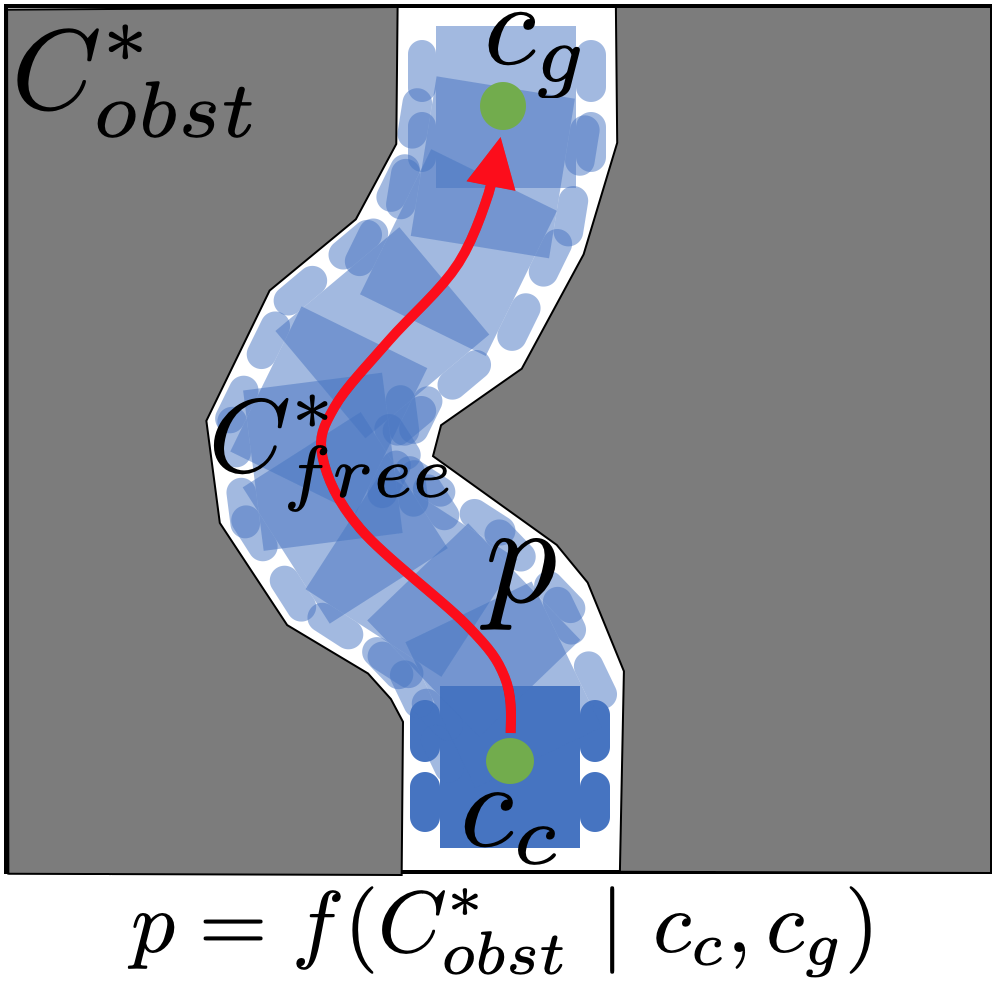}%
\label{fig::h3}}
\hspace{1pt}
\subfloat[]{\includegraphics[width=0.237\columnwidth]{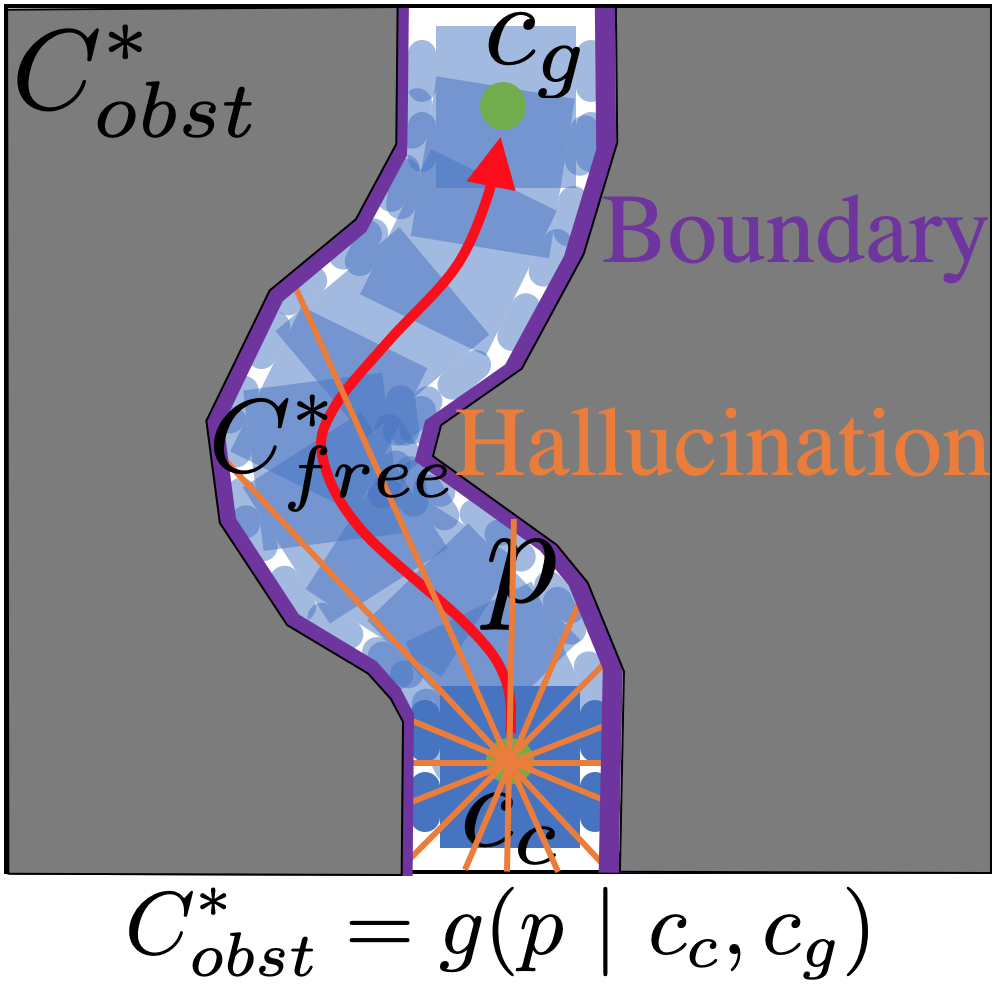}%
\label{fig::h4}}
\caption{Different unreachable sets $C_{obst}^i$ (grey) lead to the same optimal control plan $p$ (red), which generates the same trajectory (blue). But that control plan $p=f(C_{obst}^i~|~c_c, c_g)$ has a unique corresponding most constrained unreachable set $C_{obst}^* = g(p~|~c_c, c_g)$. During training, the hallucinated $C_{obst}^*$ is mapped to the robot's sensors, e.g. LiDAR range readings (orange) computed by ray casting to the boundary between $C_{obst}^*$ and $C_{free}^*$ (purple).}
\label{fig::halluci}
\vspace{-5pt}
\end{figure}

Instead of finding $f(\cdot)$, consider now its ``dual" problem, i.e., given $p$ (with $c_c$ and $c_g$), find the unreachable set $C_{obst}$ that generated that plan. Since different $C_{obst}$ can lead to the same plan, the left inverse of $f$, $f^{-1}$, is not well defined (see Figure \ref{fig::h1} and \ref{fig::h2}). However, we can instead define a similar function $g(\cdot)$ such that $C_{obst}^* = g(p~|~c_c, c_g)$, where $C_{obst}^*$ denotes the C-space's {\em most constrained} unreachable set corresponding to $p$.\footnote{\scriptsize{Technically, $c_g$ can be uniquely determined by $p$ and $c_c$, but we include it as an input to $g(\cdot)$ for notational symmetry with $f(\cdot)$.}}  
Formally, given a plan $p$, we say 
\begin{equation}
\begin{gathered}
C_{obst}^* = g(p~|~c_c, c_g) ~~\text{iff}~~\forall C_{obst}\in \mathcal{C}_{obst},\\
f^*(C_{obst}~|~c_c, c_g) = p~~\Longrightarrow~~C_{obst} \subseteq C^*_{obst},
\end{gathered}
\end{equation}
where $f^*(\cdot)$ is the optimal planner. Here, we assume every primitive control $u$ has deterministic effects, so plans have no uncertainty associated with their effects. Uncertainty will be addressed in Sections~\ref{sec::method_machine_learning} and \ref{sec::method_both}.
We denote the corresponding reachable set of $C$ as $C_{free}^* = C \setminus C_{obst}^*  $ (Figure \ref{fig::h3} and \ref{fig::h4}). We call the output of $g(\cdot)$ a \emph{hallucination} (details can be found in Section \ref{sec::implementation}), and this hallucination can be projected onto the robot's sensors. For example, for a LiDAR sensor, we perform ray casting from the sensor to the boundary between $C_{obst}^*$ and $C_{free}^*$ in order to project the hallucination onto the range readings (Figure \ref{fig::h4}). Given the hallucination $C_{obst}^*$ for $p$, the only viable (and therefore optimal) plan is $p=g^{-1}(C_{obst}^*~|~c_c, c_g)$. Note that $g(\cdot)$ is bijective and its inverse $g^{-1}(\cdot)$ is well defined. All discussed mappings are shown in Figure \ref{fig::mapping}.

\subsection{Machine Learning Solution Using Hallucination}
\label{sec::method_machine_learning}
\begin{figure}
  \centering
  \includegraphics[width=1\columnwidth]{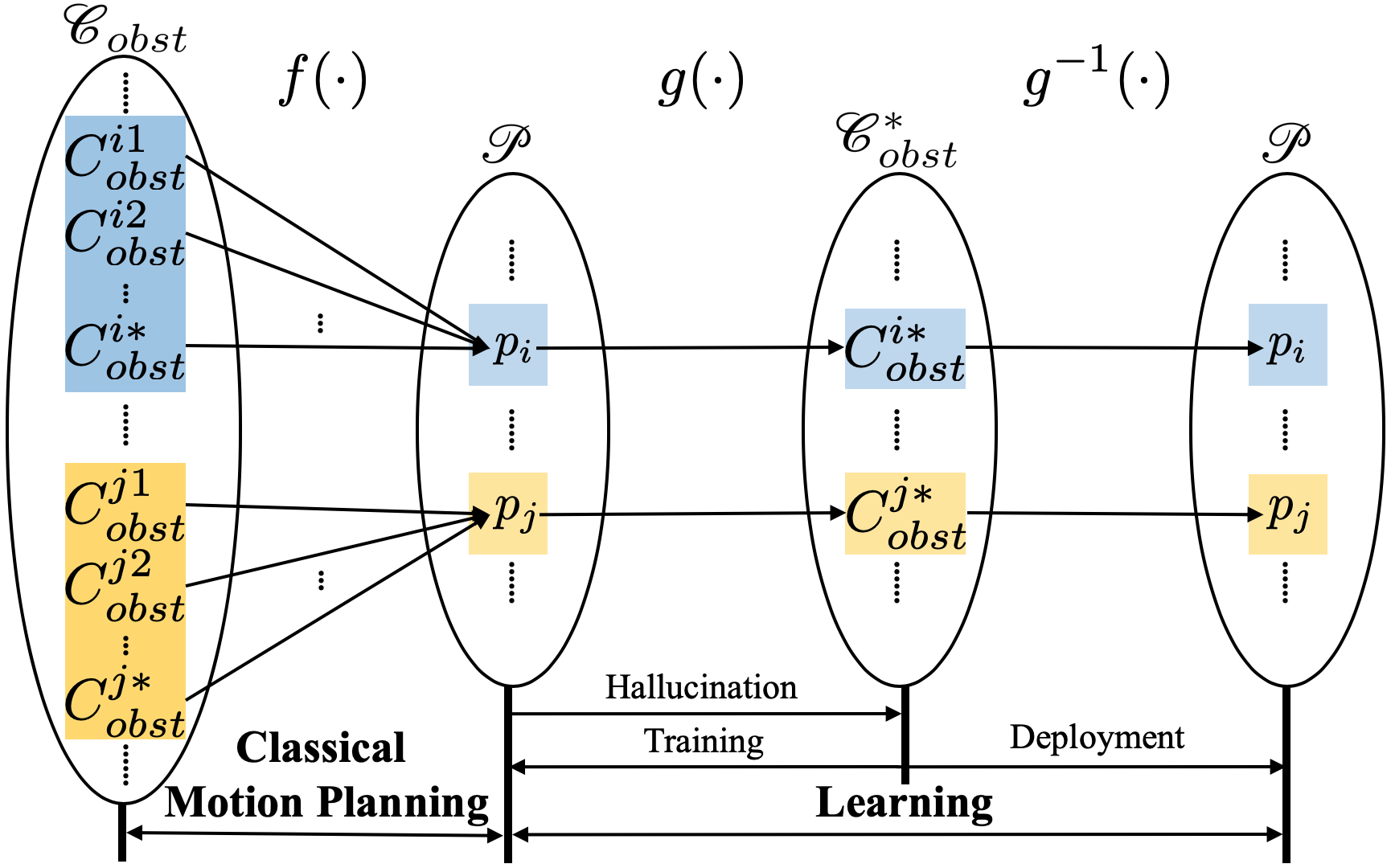}
  \caption{Classical motion planning aims at finding a function $f(\cdot)$ that maps from a larger domain of unreachable sets $\mathcal{C}_{obst}$ to plans $\mathcal{P}$. We reduce to the most constrained unreachable sets $\mathcal{C}_{obst}^*$ with hallucination $g(\cdot)$ and then use learning for $g^{-1}(\cdot)$. During deployment we hallucinate the most constrained unreachable set and predict the resulting optimal plan. }
  \label{fig::mapping}
\end{figure}

Leveraging machine learning, $g^{-1}(\cdot)$ is represented using a function approximator $g_{\theta}^{-1}(\cdot)$. Note that we aim to approximate $g_{\theta}^{-1}(\cdot)$ instead of the original $f(\cdot)$ due to the vastly different domain size: while the domain of $f(\cdot)$ is all unreachable sets $\mathcal{C}_{obst}$, $g_{\theta}^{-1}(\cdot)$'s domain is only the most constrained ones $\mathcal{C}_{obst}^*$ (Figure \ref{fig::mapping}). Solving $f(\cdot)$ demands high generalization over a large domain $\mathcal{C}_{obst}$, while a simple model $g_{\theta}^{-1}(\cdot)$ with a smaller domain $\mathcal{C}_{obst}^*$ can generalize better and robustly produce $p$. 

During training, control plans $p$ generated by a random exploration policy $\pi_{rand}$ are applied to drive the robot and the resulting sequence of robot configurations $c_i$ is recorded, where $1 \leq i \leq n$ and $n$ is the number of recorded configurations. To guarantee safety during this exploration phase even without human supervision, we conduct training in an open space without any obstacles, i.e. $C_{obst}^0 = \varnothing$ and $C_{free}^0 = C$. $C_{obst}^0$ is the most free (smallest) unreachable set, while $C_{obst}^*$ is the most constrained one. In this obstacle-free space, we use $C_{obst}^* = g(p~|~c_c, c_g)$ to generate the most constrained unreachable set, in which $c_c$ is the current robot configuration at each time step, and $c_g$ the configuration after executing $p$. The hallucinated $C_{free}^*$ can be viewed as all configurations occupied by the robot, $C_{free}^*=c_c\bigcup c_g \bigcup_{i=1}^n c_i$, and $C_{obst}^*=C\setminus C_{free}^*$. Note that we assume a deterministic world model for the hallucination and do not consider motion uncertainty. If necessary, the uncertainty can be addressed by adding an envelope around $C_{free}^*$. The hallucinated $C_{obst}^*$ is mapped to the robot's sensors: for geometric sensors, e.g. LiDAR, we hallucinate the range readings based on $C_{obst}^*$, as shown in Figure \ref{fig::h4}. In a data-driven manner, we use a function approximator, i.e. a neural network, to approximate the function $g_{\theta}^{-1}(\cdot)$ that maps from most constrained unreachable set $C_{obst}^*$ to the control plan $p$. Note that traditionally training data is collected using human demonstration or reinforcement learning, both of which are difficult in highly constrained spaces. But our inherently safe open training environment precludes the possibility of collision and allows training data to be collected in a self-supervised manner using a random exploration policy. 

\begin{figure}
\centering
\subfloat[]{\includegraphics[width=0.243\columnwidth]{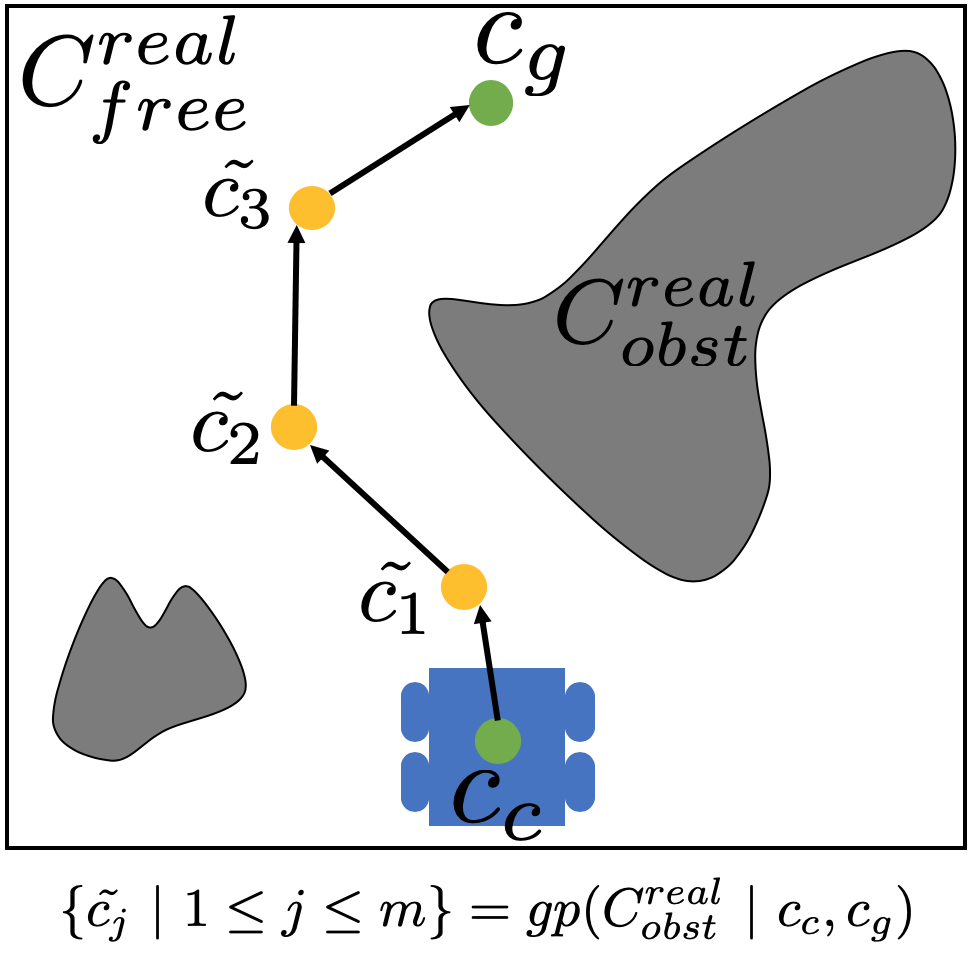}%
\label{fig::h5}}
\hspace{1pt}
\subfloat[]{\includegraphics[width=0.243\columnwidth]{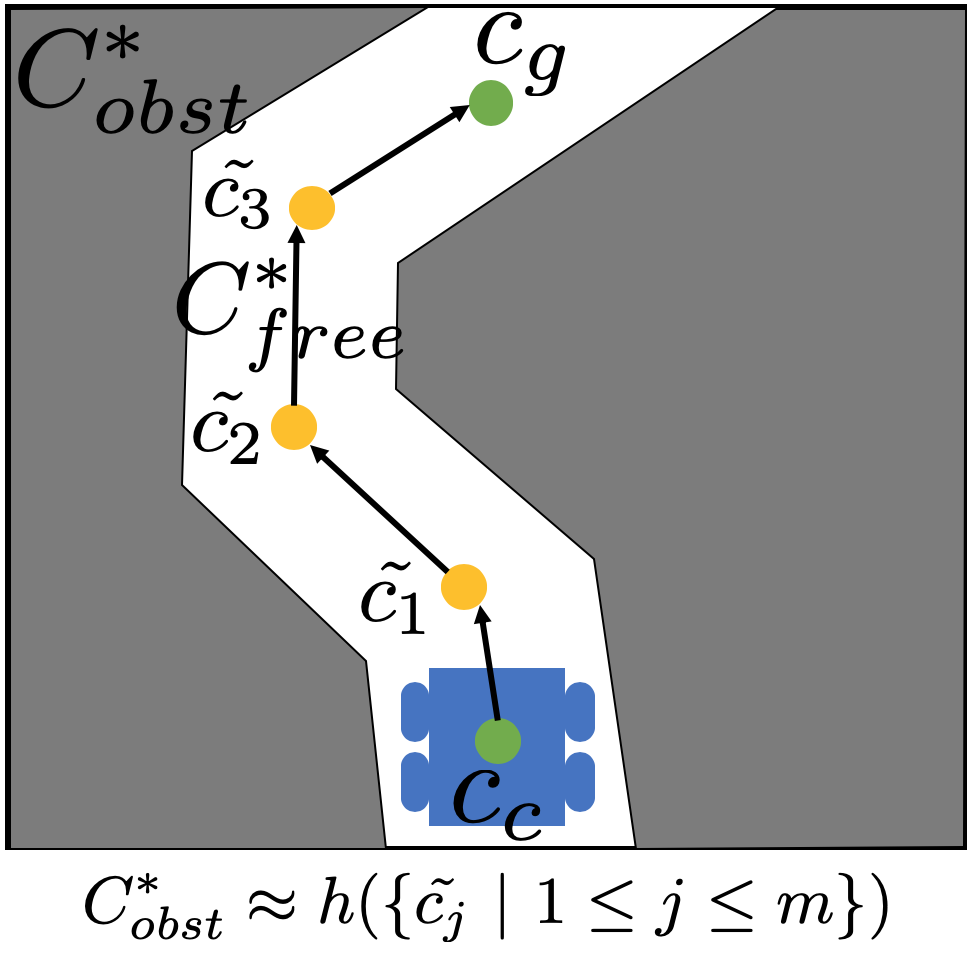}%
\label{fig::h6}}
\hspace{1pt}
\subfloat[]{\includegraphics[width=0.243\columnwidth]{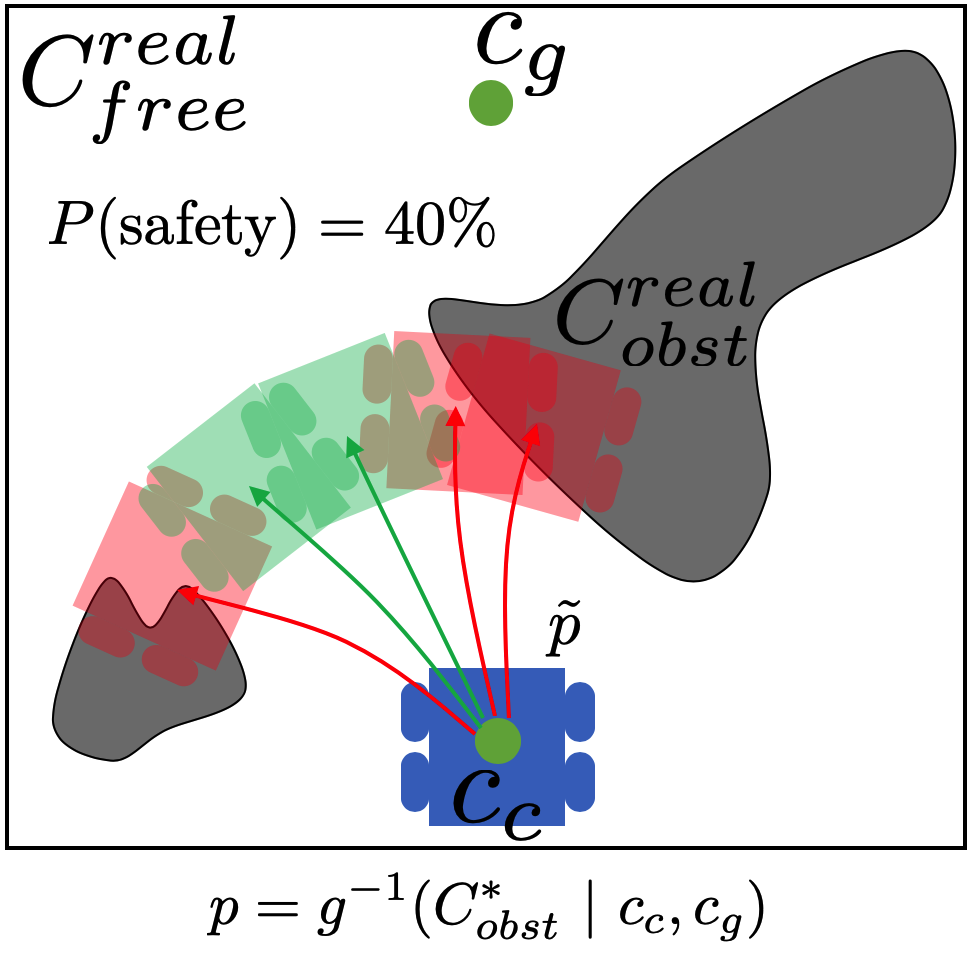}%
\label{fig::h7}}
\caption{(a) During deployment, a coarse global plan is computed quickly using real unreachable set. (b) Most constrained hallucinated unreachable set is then constructed based on the coarse global plan. (c) Before execution, safety is estimated by adding Gaussian noise to the computed plan.}
\label{fig::halluci2}
\end{figure}

During deployment, the robot uses a global planner $gp(\cdot)$ and perceives real unreachable set $C_{obst}^{real}$ to produce a coarse path: $\{\tilde{c_j}~|~1 \leq j \leq m\}=gp(C_{obst}^{real}~|~c_c, c_g)$ (Figure \ref{fig::h5}). Each configuration $\tilde{c_j}$ in this sequence can have fewer dimensions than the robot's original configuration $c_i$, and also a low resolution. For example, $c_i \in \mathbb{R}^6$ with both translational and rotational components, while $\tilde{c_j} \in \mathbb{R}^3$ with only translations. This coarse global plan can be found very quickly, with conventional search algorithms such as Dijkstra's \cite{dijkstra1959note}, A* \cite{Hart1968} or D* \cite{ferguson2006using}. Being computed in real time, it is then used to approximate the most constrained hallucinated unreachable set: $C_{obst}^* \approx h(\{\tilde{c_j}~|~1 \leq j \leq m\})$ (Figure \ref{fig::h6}). Here, $h(\cdot)$ is an online hallucination function for deployment, which maps from a sequence of approximated planned configurations to the most constrained unreachable set, instead of from a plan $p$ of actions $u_i$, as the case for $g(\cdot)$.
We hypothesize that machine learning can generalize over the difference between the codomain of $h(\cdot)$ during deployment and $g(\cdot)$ in training, and this hypothesis is validated by our experiments. Using the function $g_{\theta}^{-1}(\cdot)$ learned during the training phase, a control plan is finally computed based on the hallucinated unreachable set:

\begin{equation}
\begin{split}
p&=g_{\theta}^{-1}(C_{obst}^*~|~c_c, c_g)\\&=g_{\theta}^{-1}(h(\{\tilde{c_j}~|~1 \leq j \leq m\})~|~c_c, c_g)\\&=g_{\theta}^{-1}(h(gp(C_{obst}^{real}~|~c_c, c_g))~|~c_c, c_g).
\end{split}
\end{equation}

\subsection{Addressing Uncertainties}
\label{sec::method_both}
The LfH motion planner can only generalize well over hallucinated \emph{input} $C_{obst}^*$ that is similar to that seen in the training set, and the planning \emph{output} is not expected to assure safety during deployment. These \emph{input} differences between hallucination during deployment and hallucination in the training set and the lack of \emph{output} safety assurance during deployment motivate addressing uncertainties from both the \emph{input} and \emph{output} perspectives.

\textbf{Input Uncertainties}
Differences between hallucination during deployment and hallucination in the training set may stem from the coarse global path $\{\tilde{c_j}~|~1 \leq j \leq m\}=gp(C_{obst}^{real}~|~c_c, c_g)$ (Figure \ref{fig::halluci2}) being different from the robot trajectory in the  training set constructed by real robot trajectories (Figure \ref{fig::halluci}). 
For example, the robot trajectory in the training set may be smoother than the coarse global path computed during deployment. Another difference can arise when the global goal is behind the robot, and the coarse global planner does not consider nonholonomic motion constraints such that the planned global path may start from the current location and directly lead to somewhere behind the robot, while during training the robot has only driven forward. 
Therefore we need a \textit{pre-processing} routine that includes path smoothing and robot re-orientation to make sure the input to LfH resembles the training set.
In particular, we use smoothing within $gp(\cdot)$ to assure the coarse global path is as smooth as the robot trajectory during training so as to match the output of $h(\cdot)$ with $g(\cdot)$. Before using the LfH planner, we use a feedback controller to rotate the robot to a configuration from which the planned global path is similar to the robot trajectories in the training set (see Section \ref{sec::implementation} for implementation details). 

\textbf{Output Uncertainties}
The control plan $p$ computed by the learned function approximator $g_{\theta}^{-1}(\cdot)$ is a reaction to $C_{obst}^*$, learned from the hallucinated training set. Although it serves as an initial solution to the highly constrained workspace, it lacks both the ability to adapt to the uncertainties in real workspaces and any assurance of safety. To address those two problems, we combine the learned control plan with classical Model Predictive Control (MPC) techniques. 
When executing the learned plan $p$, we assume the output uncertainty can be expressed by Gaussian noise over the nominal inputs $u_i$. Therefore, within the computed plan $p$, we sample noisy controls around the planned inputs, $\tilde{u}_i = u_i + \epsilon_i,~\epsilon_i \sim N(0, \sigma^2 I)$, and compose a perturbed plan $\tilde{p} = \{\tilde{u}_i~|~1\leq i \leq t\}$. We then use MPC to simulate the robot under these inputs. We check each resulting trajectory in the real workspace for collision and compute the percentage of trajectories that will not enter $C_{obst}^{real}$:
$P(\text{safety}) = 1-\mathbb{E}_\epsilon [collision( \tilde{p}~|~C_{obst}^{real})]$, where $\epsilon=\{\epsilon_i~|~1\leq i\leq t\}$,
as shown in Figure \ref{fig::h7}. The magnitude of the learned controls $u_i$ is then heuristically modulated, while still observing kinodynamic constraints of the robot. For example, the robot may move fast when $P(\text{safety})$ is high and slow down when it is low. Before execution, the same MPC model checks if the modulated controls will result in a collision. If so, the controls are ignored and the robot executes a pre-specified recovery behavior. The learned control plan from hallucination can therefore both adapt to real world uncertainties and assure motion safety. 

\begin{algorithm}
 \caption{LfH Pipeline}
 \begin{algorithmic}[1]
 \renewcommand{\algorithmicrequire}{\textbf{Input:}}
 \REQUIRE $\pi_{rand}$, $g(\cdot)$, $g_{\theta}^{-1}(\cdot)$, $gp(\cdot)$, $h(\cdot)$, $collision(\cdot)$, \emph{pre-processing}, \emph{recovery behavior}.
\\\hrulefill
  \STATE // \textbf{Training}
  \STATE collect motion plans $\{p, c_c, c_g\}$ from $\pi_{rand}$ in free space and form training data $\mathcal{D}_{train} = \{C^*_{obst}, p, c_c, c_g\}$, where $C^*_{obst} = g(p~|~c_c, c_g)$
  \STATE train $g^{-1}_{\theta}(\cdot)$ with $\mathcal{D}_{train}$ by minimizing the error $\mathbb{E}_{(C^*_{obst}, p, c_c, c_g) \sim \mathcal{D}_{train}} \big[ \ell(p, g^{-1}_\theta(C^*_{obst}~|~c_c, c_g))\big]$
\\\hrulefill
  \STATE // \textbf{Deployment} (each time step)
  \STATE receive $C_{obst}^{real}, c_c, c_g$
  \STATE coarse $\{\tilde{c_j}~|~1 \leq j \leq m\}=gp(C_{obst}^{real}~|~c_c, c_g)$
  \IF {$\{\tilde{c_j}\}$ does NOT resemble training set}
  	\RETURN $p$ with \emph{pre-processing} routine
  \ENDIF 
  \STATE hallucinate $C_{obst}^* \approx h(\{\tilde{c_j}~|~1 \leq j \leq m\})$
  \STATE plan $p=\{u_i~|~1\leq i \leq t\}=g_{\theta}^{-1}(C_{obst}^*~|~c_c, c_g)$
  \STATE add noise $\epsilon = \{\epsilon_i\}$ to $p$, $\tilde{p}=\{\tilde{u}_i~|~1\leq i \leq t\}$
  \STATE modulate $p$ based on \\$P(\text{safety}) = 1-\mathbb{E}_\epsilon [collision( \tilde{p}~|~C_{obst}^{real})]$
  \IF {$p$ is NOT safe}
    \RETURN $p$ with \emph{recovery behavior}
  \ENDIF
 \RETURN $p$
 \end{algorithmic}
 \label{alg::lfh}
 \end{algorithm}

\textbf{LfH Pipeline}
The LfH motion planning pipeline is shown in Algorithm \ref{alg::lfh}. In lines 2 -- 3, training data is collected via a random exploration policy in obstacle-free space and $g_{\theta}^{-1}(\cdot)$ is trained on hallucinated data. At each step during deployment, line 6 computes a coarse global path from $gp(\cdot)$. Lines 7 -- 9 correspond to the pre-processing routine based on a feedback controller to address input uncertainties. Line 10 hallucinates and line 11 produces motion plans. Lines 12 -- 16 address the output uncertainties.

\section{EXPERIMENTS}
\label{sec::result}
In this section, LfH is implemented on a ground robot in simulation and the real-world. We hypothesize that LfH can achieve safer, faster, and smoother navigation than existing classical and learning approaches in highly constrained spaces. We first present our implementation of LfH on the robot. Baseline methods to compare LfH against are then described, and we provide quantitative experimental results in 300 simulation environments and a real-world controlled test course, and qualitative results in natural outdoor/indoor environments. 

\subsection{Implementation}
\label{sec::implementation}
A Clearpath Jackal, a differential drive four-wheeled ground robot, is used as a test platform for the LfH motion planner. Jackal's nonholonomic constraints increase the difficulty of maneuvering in highly constrained spaces. Taking advantage of the widely-used Robot Operating System (ROS) \texttt{move\textunderscore base} navigation stack \cite{ros_move_base}, we replace its DWA local planner with our LfH pipeline (Algorithm \ref{alg::lfh}), and use the same high-level global planner (Dijkstra's algorithm).
The global planner assumes unknown regions are free and replans when obstacles are perceived. The local environment is assumed to be known to the local planner. 
Input $C_{obst}^{real}$ is instantiated as a 720-dimensional 2D laser scan with a limited 1m range. Other types of metric/geometric perception, such as depth images, can also be used to instantiate $C_{obst}^{real}$. Planning in a robot-centric view, $c_c$ is the origin and $c_g$ is a waypoint 1m away on the global path (orientation is ignored for simplicity). 

\begin{figure}
  \centering
  \includegraphics[width=1\columnwidth]{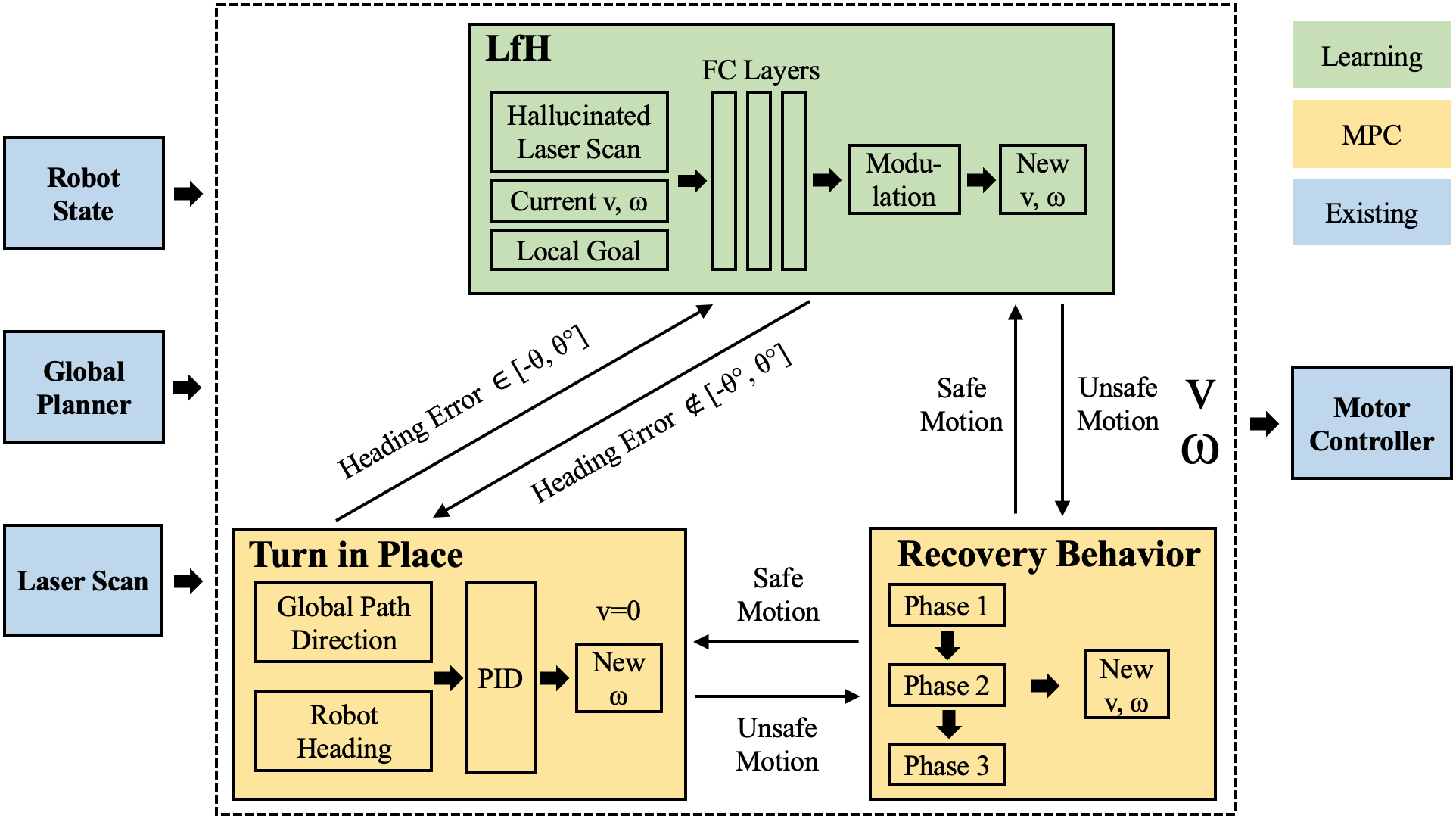}
  \caption{Finite State Machine of the LfH Implementation}
  \label{fig::fsm}
  \vspace{-5pt}
\end{figure}
For training, data is collected in a self-supervised manner with a Jackal robot in real-time simulation (line 2 in Algorithm \ref{alg::lfh}). The planning horizon $t$ is set to 1, i.e. only a single action $u_1 = \left(v_1, \omega_1\right)$ (linear and angular velocity) is produced, for faster computation and better accuracy. The random exploring policy $\pi_{rand}$ is simulated by a human operator randomly pushing an Xbox joystick\footnote{A random exploration policy implemented later also works well.}, with $v$ bounded in $[0, 0.4]$m/s and $\omega$ in $[-1.4, 1.4]$rad/s. This $\pi_{rand}$ could be easily replaced by a random walk exploration policy. Linear and angular velocities are randomly applied to the robot and perform a large variety of different maneuvers. We record the control inputs ($v$ and $\omega$) as training labels. For simplicity, we directly record robot configurations ($x$, $y$, and $\psi$) from simulation ground truth, instead of computing them based on $v$ and $\omega$, to extract local goals and to hallucinate LiDAR as training input. The inherent safety provided by the collision-free open training environment allows completely self-supervised learning of a rich variety of motions. We speculate that even when training in an open space in the real world, safety can be assured by a collision avoidance policy to drive the robot back into the middle of the open space when it comes close to the environment boundary. We find the model (three-layer neural network, with 256 hidden neurons and ReLU activation for each layer) learned from 
the less than four minutes simulation in real time can easily generalize to the real world. Training with the four-minute data takes less than one minute on an Intel Core i9-9980HK CPU, indicating high computational efficiency (line 3). 

For deployment, we implement a Finite State Machine, shown in Figure \ref{fig::fsm}. 
For line 6 in Algorithm \ref{alg::lfh}, $\{\tilde{c_j}~|~1 \leq j \leq m\}$ is smoothed by a Savitzky–Golay filter \cite{krishnan2012selection} on the global path. In general, path smoothing may lead to an invalid plan, but we do not observe such an effect from smoothing the coarse global path planned by Dijkstra's algorithm in \texttt{move\textunderscore base}. For lines 7 -- 9, a PID controller in the \emph{pre-processing} routine rotates the robot in place to address out-of-distribution scenarios unseen in the training set (when the angle between the current robot heading and the current tangential direction of the global path falls out of the range $[-30^{\circ}, 30^{\circ}]$). Otherwise, the LfH module takes control. For line 10, $C_{obst}^*$ is constructed as all configurations that are slightly more than the robot's half-width away from the smoothed global path. We use ray casting to generate hallucinated LiDAR input. For safety, we take the minimum value between the hallucinated and real laser scan. 
Concatenated with the current linear and angular velocity and the local goal $c_g$ taken from the global plan, the hallucinated LiDAR is fed into a neural controller ($g_{\theta}^{-1}(\cdot)$) and a plan with planning horizon $t=1$ (one command of linear and angular velocity) is produced (line 11). We add Gaussian noise of zero mean and 10\% standard deviation to the produced $v$ and $\omega$ in line 12. The controls are modulated by the safety estimation by a MPC collision checker in line 13:
\[
e^{w_1-w_2(1-P(\text{safety}))}\cdot \{v, \omega\},
\]
where the weights for the speed modulation $w_1=0.4$ and $w_2=1.0$ correspond to roughly 50\% - 150\% modulation. In order to prevent noisy $\omega$ being amplified by the modulation in safe spaces, we suppress any $\omega < 0.04$rad/s to $0$. 
For the \emph{recovery behavior} routine in lines 14 -- 16 when a collision is detected, the robot starts a three-phase recovery behavior designed by hand. If the first (decreasing $v$ and increasing $\omega$ iteratively) and second (increasing negated $v$ and original $\omega$ iteratively) phase are still unsafe, the robot drives back slowly in the third phase. 

\subsection{Results}
\begin{figure}
  \centering
  \includegraphics[width=1\columnwidth]{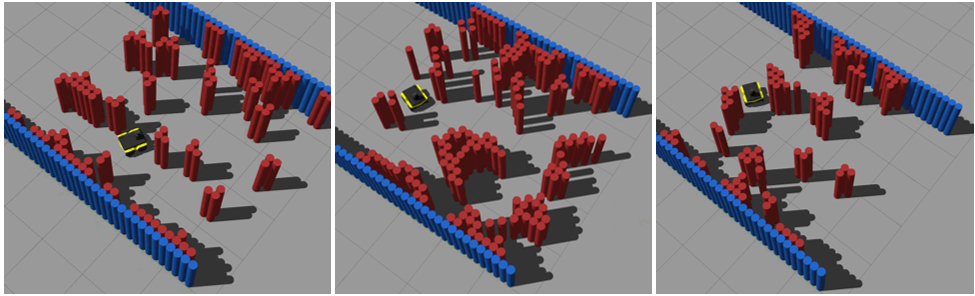}
  \caption{Three Example Simulation Environments}
  \label{fig::simulated_envs}
\end{figure}

\begin{figure}
  \centering
  \includegraphics[width=1\columnwidth]{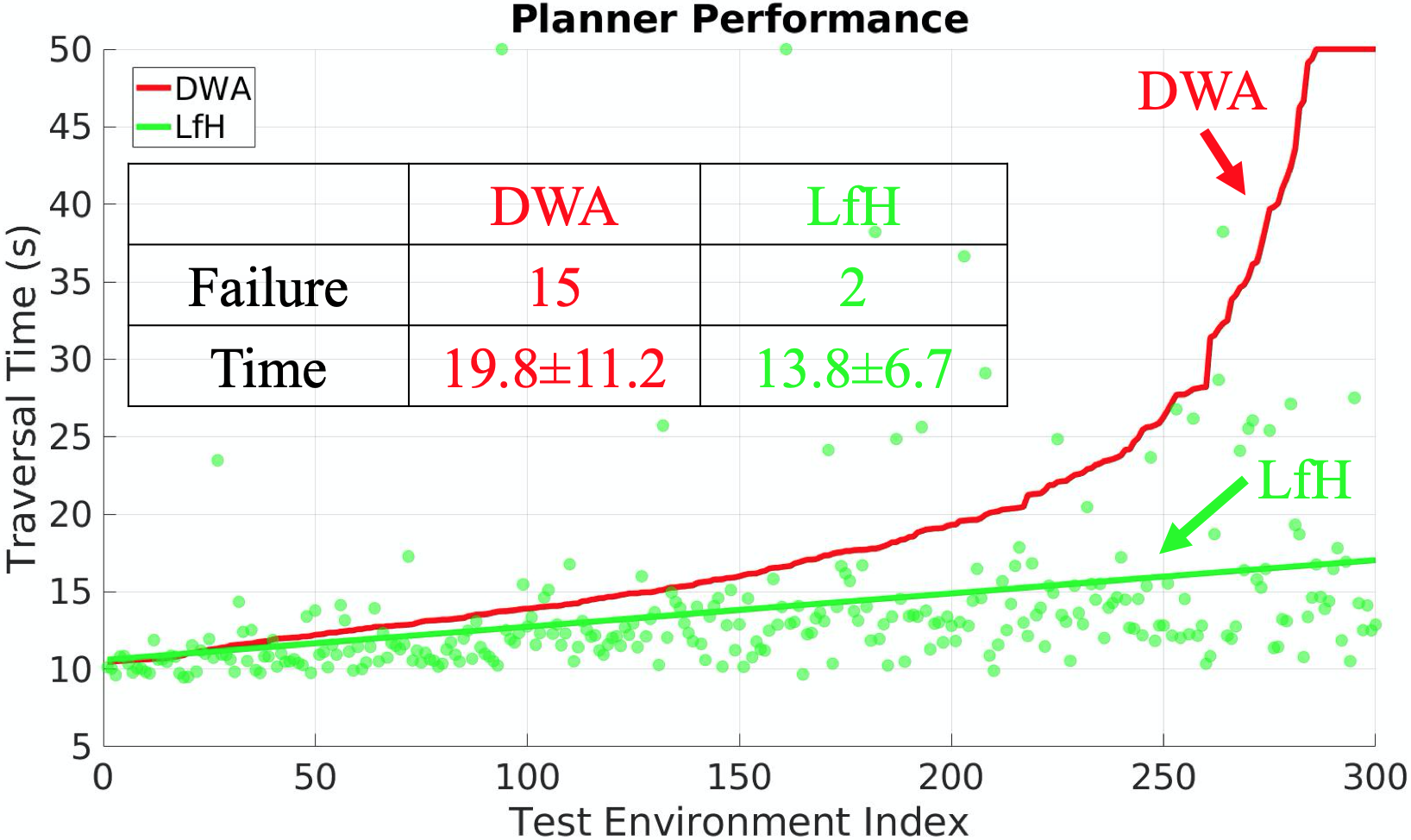}
  \caption{Simulation Results: Number of failure environments, average traversal time of all trials of DWA and LfH (table), and their performance in the 300 individual BARN environments (curve, averaged over three trials each environment).}
  \label{fig::dwa_lfh}
\end{figure}

We first test LfH against the sampling-based DWA planner of \texttt{move\textunderscore base} in the Benchmark Autonomous Robot Navigation (BARN) dataset~\cite{perille2020benchmarking}, which uses traversal time as a metric to benchmark navigation performance. BARN is composed of 300 navigation environments randomly generated using Cellular Automata. 
The DWA planner is a widely used representative classical navigation approach. 
The default parameters recommended by the robot manufacturer\footnote{\url{https://github.com/jackal/jackal/blob/melodic-devel/jackal_navigation/params/base_local_planner_params.yaml}} are used. We acknowledge that if re-tuned for each environment DWA can achieve better performance. But it is not practical to do so for all 300 environments, neither with expert knowledge, nor with state-of-the-art learning methods~\cite{pfeiffer2017perception, xiao2020appld}. In each of the 300 environments, the robot navigates three trials between a specified start and goal location without a map, using the default DWA and the LfH planner. This results in 1800 total trials and the maximum traversal time of each trial is capped at 50s (more than 50s is defined as failure). Three example simulation environments are shown in Figure \ref{fig::simulated_envs}. 
We order the environments using the average traversal time of DWA (red) as an indicator of difficulty/constrained level, and also show the average LfH performance (green) in Figure \ref{fig::dwa_lfh}: although navigating only at a low 0.5m/s speed, the default DWA still fails to sample feasible actions in many scenarios, causing back and forth motion or getting stuck. In most cases, LfH achieves equal or faster traversal times than DWA, and the green line fitted to the green dots indicates LfH is less sensitive to difficult/constrained environments than DWA. 

The LfH controller is then tested in a real-world highly constrained obstacle course with minimum clearance roughly 1.3$\times$ the robot's footprint (see Figure \ref{fig::gdc}), whose difficulty is evident based on the 100\% failure rate of the robot's default planner.  
The robot needs to perform agile maneuvers to navigate through this environment without a map. 
We compare the LfH controller with three baseline local planners. First, we compare to the default DWA again, as the classical sampling-based motion planner. Second, a machine learning approach, similar to the Behavior Cloning (BC) work presented by Pfeiffer {\em et al.} \cite{pfeiffer2017perception}, replaces the green LfH module in Figure \ref{fig::fsm} and maintains the yellow Turn in Place and Recovery Behavior modules to study the effect of hallucinated learning. We use the same neural network architecture as LfH, but with realistic instead of hallucinated LiDAR input. Note that the original BC work used DWA as the expert, so the learned performance is upper bounded by DWA. We directly compare to their expert and further provide the BC framework with a better expert: human expert demonstration in the same deployment environment. Third, as a recent improvement upon classical planners, APPLD \cite{xiao2020appld} can fine-tune DWA parameters based on different navigation contexts to improve navigation. Therefore, we compare to APPLD-DWA as an upper bound of classical approaches. In particular, one of the authors with extensive Jackal teleoperation experience provides a demonstration in the same deployment environment for BC and APPLD. The demonstrator aims at quickly traversing the course in a safe manner. 
APPLD generates four sets of planner parameters to adapt to different regions of the obstacle course. 
Note that with demonstration in the same environment, we allow BC and APPLD to fit to the deployment environment, while LfH is not given any such training data. 
All approaches use exactly the same hardware. 

\begin{table}
\centering
\caption{Quantitative Results}
\begin{tabular}{ccccc}
\toprule
          & DWA & BC & APPLD &  \textbf{Hallucination} \\ 
\midrule
\textcolor{green}{Success}   & \textcolor{green}{0} & \textcolor{green}{2} & \textcolor{green}{8} &  \textbf{\textcolor{green}{9}} \\
\hdashline
\textcolor{yellow}{Collision} & \textcolor{yellow}{0} & \textcolor{yellow}{5} & \textcolor{yellow}{2} &  \textbf{\textcolor{yellow}{1}} \\
\textcolor{red}{Failure}   & \textcolor{red}{10} & \textcolor{red}{3} & \textcolor{red}{0} & \textbf{\textcolor{red}{0}} \\
\hdashline   
Time      &  $\infty$ & 53.9$\pm$6.1  & 74.1$\pm$2.8 & \textbf{45.8$\pm$1.4} \\
\bottomrule
\end{tabular}
\label{tab::results}
\end{table}

Ten trials for each planner are executed in the obstacle course. The results of the four methods are summarized in Table \ref{tab::results}. We define ``Success" as navigating through the course without any collision. Non-terminal ``Collision"s are recorded and we allow the robot to keep navigating to the goal. ``Failure" is when the robot fails to navigate through the course to reach the destination, e.g. getting stuck at a bottle neck or knocking down the obstacle course. We compute average traversal time based on the ``Success" and ``Collision" trials. For default DWA, all ten trials get stuck mostly because the default planner cannot sample viable velocity commands in such a highly constrained environment. BC does not finish three out of ten trials, mostly due to knocking down the course, and in five other trials, the robot collides with the obstacles. The safety check fails to prevent some collisions since the close distance to the obstacle is smaller than the LiDAR's minimum range. The other two trials are successful, but we observe very jittery motion. Using four sets of parameters learned in eight hours from human demonstration, APPLD successfully navigates Jackal through the obstacle course without collision in eight trials, and causes one gentle collision in each of the other two trials. The average traversal time is longer than the demonstration (67.8s). The LfH controller succeeds in nine out of ten trials with a faster average speed and smaller variance. The motion is qualitatively smoother than other methods. One gentle collision with an obstacle happens in one trial. 

\begin{figure}
  \centering
  \includegraphics[width=1\columnwidth]{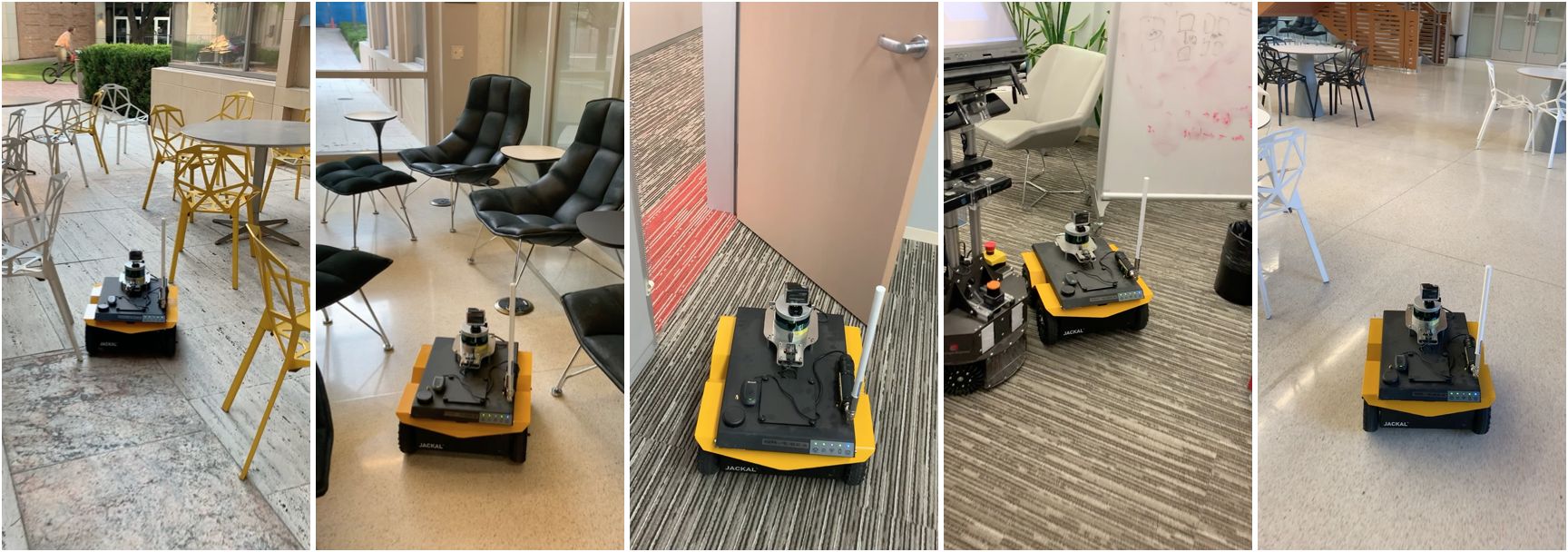}
  \caption{Qualitative Results: Jackal navigates with the LfH local planner in outdoor and indoor environments with highly cluttered natural objects and tight clearances. (\url{https://www.youtube.com/watch?v=AE-KgxJS-iE})}
  \label{fig::natural}
  \vspace{-12pt}
\end{figure}

Qualitatively, we also test the LfH local planner in both outdoor and indoor natural environments. It is able to successfully navigate through highly constrained spaces cluttered with everyday objects, including tables, chairs, doors, white boards, trash cans, etc. In relatively open space, LfH can also enable smooth and fast navigation. Note the robot has \emph{never} seen any of those environments, nor even a single real obstacle, during training, and does not require human supervision. 
Admittedly, better demonstrations or extensive engineering targeted at the deployment environment could enable conventional learning or classical approaches to match or even exceed LfH’s performance.  The strength of LfH is in its performance without such demonstrations or engineering, and in its ability to generalize to unseen deployment scenarios (as shown in Figures \ref{fig::gdc}, \ref{fig::simulated_envs}, \ref{fig::dwa_lfh}, and \ref{fig::natural}).



\section{CONCLUSION}
\label{sec::conclusion}
This paper introduces the novel LfH technique to address motion planning (i.e., navigation) in highly constrained spaces. For robotics, the LfH method addresses the difficulty in planning motion when obstacle space occupies the majority of the surrounding C-space, which usually causes an increasing demand on sampling density and therefore computation using classical approaches. Seeking help from a data-driven perspective, we hallucinate the most constrained workspace that allows the same effective maneuvers in open space to the robot perception and learn a mapping from the hallucinated workspace to that optimal control input. For machine learning, the proposed LfH method provides a self-supervised training approach and largely improves sample efficiency, compared to  traditional IL and RL. To combine the benefits of both sides, our local planner estimates motion safety with MPC and enables agile maneuvers in highly constrained spaces with learning. In simulated and physical robot experiments, LfH outperforms classical sampling-based method with default and even with dynamically fine-tuned parameters, and also imitation learning deployed in the identical training environment. An interesting direction for future work is to extended LfH beyond 2D ground navigation, e.g. toward 3D aerial navigation or manipulation with higher degrees of freedom. In this case, new hallucination techniques beyond simple 2D ray casting are needed, e.g. for 3D LiDAR or depth camera. Another interesting direction is to investigate hallucination techniques that not only hallucinate the most constrained partition $C = C_{obst}^* \cup C_{free}^*$, but also any partition $C = C_{obst}^i \cup C_{free}^i$ in between, for which the motion plan $p$ is still optimal. In that case, hallucination during deployment is no longer necessary.


\section*{ACKNOWLEDGMENT}
This work has taken place in the Learning Agents Research Group (LARG) at UT Austin.  LARG research is supported in part by NSF (CPS-1739964, IIS-1724157, NRI-1925082), ONR (N00014-18-2243), FLI (RFP2-000), ARO (W911NF-19-2-0333), DARPA, Lockheed Martin, GM, and Bosch.  Peter Stone serves as the Executive Director of Sony AI America and receives financial compensation for this work.  The terms of this arrangement have been reviewed and approved by the University of Texas at Austin in accordance with its policy on objectivity in research.

\bibliographystyle{IEEEtran}
\bibliography{IEEEabrv,references}

\end{document}